\def\year{2019}\relax
\documentclass[letterpaper]{article} 
\usepackage{aaai19}  
\usepackage{times}  
\usepackage{helvet}  
\usepackage{courier}  
\usepackage{url}  
\usepackage{graphicx}  

\usepackage{amsmath}
\usepackage{amssymb}
\usepackage{fixmath}
\usepackage{algorithm}
\usepackage{algorithmic}
\usepackage{booktabs}

\nocopyright
\begin{document}
%
\title{Skeptical Deep Learning with Distribution Correction}
\author{Mingxiao An\textsuperscript{1}, Yongzhou Chen\textsuperscript{1}, Qi Liu\textsuperscript{1,*}, Chuanren Liu\textsuperscript{2}, Guangyi Lv\textsuperscript{1}, Fangzhao Wu\textsuperscript{3}, Jianhui Ma\textsuperscript{1,*} \\
	\textsuperscript{1}Anhui Province Key Laboratory of Big Data Analysis and Application, School of Computer Science and Technology, \\University of Science and Technology of China \\ \{amnx, cyz, gylv\}@mail.ustc.edu.cn, \{qiliuql,jianhui\}@ustc.edu.cn \\ \textsuperscript{2}Drexel University, chuanren.liu@drexel.edu \\ \textsuperscript{3} Microsoft Research Asia, China, fangzwu@microsoft.com}
\maketitle
\begin{abstract}
Recently deep neural networks have been successfully used for various classification tasks, especially for problems with massive perfectly labeled training data. 
However, it is often costly to have large-scale credible labels in real-world applications. 
One solution is to make supervised learning robust with imperfectly labeled input. 
In this paper, we develop a distribution correction approach that allows deep neural networks to avoid overfitting imperfect training data.
Specifically, we treat the noisy input as samples from an incorrect distribution, which will be automatically corrected during our training process. 
We test our approach on several classification datasets with elaborately generated noisy labels.
The results show significantly higher prediction and recovery accuracy with our approach compared to alternative methods. 

\end{abstract}

\newcommand{\x}{\mathbold{x}}
\newcommand{\ty}{\tilde{y}}
\newcommand{\hy}{\hat{y}}
\newcommand{\param}{\mathbold{\theta}}

\section{Introduction}
Generally, label noise comes from the stochastic process that the labels subject to before being presented to the learning algorithm \cite{lfn88}.
Typically, the noise is brought to the dataset during annotating.
In recent years, deep neural networks have achieved great success in classification tasks, especially those with large perfectly labeled datasets. 
In some applications, however, it is very costly to annotate such large datasets by expert level annotators. 
Accepting amateur annotators or crowdsourcing are good solutions, but those labels should be less credible, which result in label noise. 
Moreover, some untrusted annotators may label maliciously \cite{poi18}, which is a tricky problem.

We aim for the cases that no any other prior information has been obtained except the noisy-labeled dataset. 
Neither a set of clean labels nor the model of label noise is available. 
The problem is, since their validation sets are also with noisy labels in these cases, tuning can encounter an obstacle.
Therefore, it is important for a solution to gain the confidence of the users.
The method should be adaptive in order to face the various type of label noise.
Also, methods with both good theoretical and empirical result will gain more application value.
Our goal is to find the best method that meets the criteria above.

There are several challenges here.
First, we need to find a framework that accommodates most kinds of noise types.
Second, we should derive a practical method that is robust to various types of label noise.
Also, we need massive datasets with label noise that is difficult for a deep neural network.
We should figure out how to generate those confusing labels from existed perfect datasets.

In this paper, we first introduce an approach named distribution correction.
We assume that the noisy dataset is sampled from an incorrect distribution.
Then, we derive how the expectation of values in the correct distribution can be represented in the noisy distribution.
Also, We use this approach to explain the forward loss correction \cite{lca17}.
We implement the distribution correction by skeptical learning, where we substitute the correct distribution by the expressions with the model's predictions.

Although deep neural networks are usually more tolerant to the massive label noise \cite{rol17} compared to other machine learning algorithms such as SVMs \cite{net10}, many previous works \cite{unh15,lfm15,nal16,rob17,lca17,tcn14} have shown that the mere logistic loss is not the best option.
Typically, they either changed the loss function, or added dynamic processes to each label.
People can also adopt heuristic solutions such as \(\lambda\)-trick \cite{rel00} or \(\alpha\)-bound \cite{bou00}, which use the times of prediction failures in the training process as a measure of the confidence to each data entry. 
We believe these heuristic solutions are based on the generalization ability of a model and force the model to be more consistent with itself. 
However, when it comes to complex classification tasks of multiple labels, it usually takes many iterations to converge to the extent that we have enough confidence in its predictions.
As the dataset we have is polluted, we usually do not know when the predictions have become referable, so the solution should be adaptive. 
Finally, our solution combines these two aspects into an adaptive training process.

We also propose a solution to generate a noisy-labeled dataset that is confusing to a model from an existed dataset, by sampling the incorrect label from the predictions of a trained deep neural networks. 
The type of label noise can be categorized as a completely random, a label relevant, and both label and feature dependent random process \cite{sur14}. 
Usually, The latter two types better describe the noise in reality, while they are difficult to be generated manually.
Indeed, researchers can collect actual noisy labels \cite{crf17}, but it is still costly to collect enough such datasets to do massive tests.
Our solution generates noisy datasets of high quality in large quantities, which can make the empirical experiments more convincing.

\section{Related Works}
In this section, we introduce the background of label noise study.
For the cases that we have no any information about the noise, forword loss correction should gain notice.
We use the model's predictions during training, which can be treated as proxy labels in self-ensembling.

\paragraph{Label Noise Robustness and Tolerance}
Many previous works on label noise robustness or tolerance learning approaches are well described by \cite{sur14}. 
As for traditional problems, \cite{unh15} has proved that the unhinged loss is a convex loss that is theoretically robust to symmetric label noise on binary classification.
In recent years, label noise robustness algorithms on deep neural networks have been successful in dealing with datasets when we have a small set with clean labels \cite{crf17}\cite{unr16}. 

\paragraph{Forward Loss Correction}
The method proposed by \cite{tcn14} adds a linear layer between softmax and cross-entropy loss function. 
The layer was then identified as a transition matrix by \cite{lca17}, who name this method forward correction. They have also proposed the backward loss correction, which, they believe, has a better theoretical property compared to that of forward correction. However, the experiments result in that paper and \cite{crf17} show that the models trained with forward loss correction have higher accuracy on ground truth labels. 
We will explain why the forward one is superior according to the distribution correction model.

\paragraph{Self-ensembling or Self-training}
It is widely used in semi-supervised learning problems that a model can learn from its own predictions under different configuration \cite{pim16} or outputs by a mean teacher whose weights is the moving average that of the model \cite{mea17}. 
These labels can be regarded as proxy labels.
The success of these models in semi-supervised learning indicates that the predictions during training are at least referable. 
We think these semi-supervised approaches worth a notice, and \cite{boo14} has tried to solve the potential contradictions between proxy labels and noisy labels.

\section{Preliminaries}
We start with fixing notations.
In the multi-class classification problems with label noise, the training dataset \(\tilde{D}=\{(\x_1, \ty_1), (\x_2, \ty_2), \cdots, (\x_N, \ty_N)\}\) is the noisy dataset.
Each \(\x_i\) belongs to the feature space \(\mathcal{X}\), while each \(y_i\) belongs to the label space \(\mathcal{Y}\).
The labels in the label space \(\mathcal{Y}\) are mutually exclusive, so there are altogether \(|\mathcal{Y}|\) different labels.
We assume that the corresponding clean dataset \(D=\{(\x_1, y_1), (\x_2, y_2), \cdots, (\x_N, y_N)\}\) exists.
For each entry \((\x_i, \ty_i)\) in \(\tilde{D}\), there is a corresponding entry with true label \(y_i\) in \(D\).
The corresponding labels may or may not be equal. 

The goal of the label noise robust learning is to achieve the maximum accuracy on the test dataset, under the restriction that only the noisy training dataset is available before evaluation.
Beside the accuracy on the test dataset and the clean training dataset \(D\), we also evaluate the recover performance by the precision and recall.
Those metrics are defined according to those in the label noise cleansing task by \cite{eli03}.
We take the trained model's prediction as the recovered dataset \(\hat{D}=\{(\x_1, \hy_1), (\x_2, \hy_2), \cdots, (\x_N, \hy_N)\}\).
With \(i \in \{1, 2, \cdots, N\}\), the metrics are defined as:

\begin{align}
	\begin{split}
	\label{equ:precision}
	\mathrm{precision} &= P(\hy_i = y_i | \hy_i \neq \ty_i)
	\end{split}
	\\
	\begin{split}
	\label{equ:recall}
	\mathrm{recall} &=  P(\hy_i = y_i | y_i \neq \ty_i).
	\end{split}
\end{align}

We define the portion of the wrong label pairs between the noisy and clean datasets as the label noise rate \(p_e\).
We do not consider the situation that either \(p(\ty_i = y | y_i = y) \ge 0.5\) or \(p(y_i = y | \ty_i = y)\) is satisfied for any \(y \in \mathcal{Y}\).
It makes the label flip possible during training, when the major data entries of a label in the clean dataset are classified into the same wrong category.
Even though the model could suffer such unbalanced noise on some occasions, it is beyond our consideration.

\begin{figure}
\centering
\includegraphics[width=0.45\textwidth]{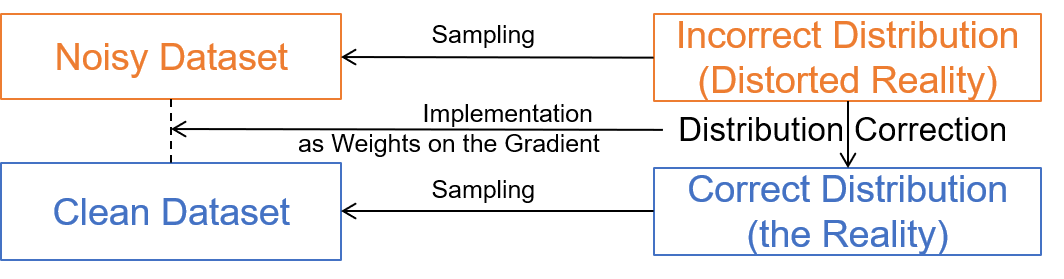}
\caption{The overview of distribution correction.}
\label{fig:overview}
\end{figure}

The distribution correction approach assumes that each dataset is sampled from a corresponding generative model \(G\).
As illustrated in Fig.(\ref{fig:overview}), each generative model represent a distribution of instances, which is consistent with the reality or a distorted reality.
Datasets are considered as samples from these models.
In this way, we first investigate how the generative model can be corrected, and then try to implement the correction on the learning process.
We note the collection of all the possible generative models as \(\mathcal{G}\).
The corresponding models for the noisy and clean datasets are \(G_{\tilde{D}}\) and \(G_D\).
These two models have the maximum likelihood to sample the datasets: \(G_{\tilde{D}} = \mathop{\mathrm{argmax}}\limits_{G \in \mathcal{G}} \mathcal{L}(\tilde{D} | G), G_D = \mathop{\mathrm{argmax}}\limits_{G \in \mathcal{G}} \mathcal{L}(D | G)\).
\(G(\x, y)\) represents the event that \((\x, y)\) is sampled from \(G\).
Let \(G(\x) = \mathop{\cup}\limits_{y \in \mathcal{Y}} G(\x, y)\), and \(p(G(\x))\) is the marginal distribution of the features.
Whatever dataset, it should sample each feature in a constant distribution.
Thus, for every two models \(G_1, G_2 \in \mathcal{G}\) and any \(\x\), we have \(G_1(\x) \equiv G_2(\x)\).
We also define \(G(y) = \mathop{\cup}\limits_{x \in \mathcal{X}} G(\x, y)\).
\(G(\x, y)|G(\x)\) is the event that \((\x, y)\) is sampled from \(G\) given that \(\x\) is the feature of the sample.
Therefore, the distribution of the label for each feature is \(p(G(\x, y) | G(\x))\), which meets the following rules Eq. (\ref{equ:distribution}).
Similarly, the event \(G(\x, y) | G'(\x, y')\) and \(G(y) | G'(y')\) satisfy the Eq. (\ref{equ:total condition}) and Eq. (\ref{equ:partial condition}).

\begin{align}
	\begin{split}
		\label{equ:distribution}
		p(G(y) | G(\x)) = \frac{p(G(\x, y))}{\sum\limits_{y' \in \mathcal{Y}} p(G(\x, y'))}
	\end{split}
	\\
	\begin{split}
		\label{equ:total condition}
		p(G(\x, y)|G'(\x, y')) = \frac{p(G(\x, y) \cup G'(\x, y'))}{p(G'(\x, y'))}
	\end{split}
	\\
	\begin{split}
		\label{equ:partial condition}
		p(G(y)|G'(y')) = \frac{\sum\limits_{\x \in \mathcal{X}} p(G(\x, y) \cup G'(\x, y'))}{\sum\limits_{\x \in \mathcal{X}} p(G'(\x, y'))}
	\end{split}
\end{align}

Usually, deep neural networks are optimized by their first moment of the first gradient on the loss function.
Sometimes, the second moment of the gradient is used especially in adaptive optimizing algorithms such as Adam \cite{ada14}. 
Therefore, the equivalence of the first gradient would result in the equivalent model predictions for deep neural networks in most cases.
The first moment of the gradient can be represented in two ways: the average of the gradient for each entry in the dataset, and the expectation of the gradient in the distribution given by the corresponding generative model.
For example, these two expressions in Eq. (\ref{equ:gradient}) are equivalent, where \(\param\) is the parameters of the networks for training.

\begin{align}
	\begin{split}
		\label{equ:gradient}
		& \frac{1}{N}\sum\limits_{i=1}^N \nabla_{\param} \log(p(y_i | \x_i; \param)) \\
		& \Longleftrightarrow \sum\limits_{(\x, y) \in \mathcal{X} \times \mathcal{Y}} p(G(\x, y)) \nabla_{\param} \log(p(y | \x; \param)) \\
		& = \mathbb{E}_{(\x, y) \sim p(G(\x, y))} \nabla_{\param} \log(p(y | \x; \param))
	\end{split}
\end{align}

\section{Distribution Correction}
In this section, we first introduce the distribution correction approach in two forms: posterior correction and conditional correction.
Then, we explain why the forward loss correction method can tolerate label noise according to the approach. 
We notice that good quality of the estimation of \(p(G_D(\x, \ty)|G_{\tilde{D}}(\x, y))\) is the key to a successful correction.
For this reason, we propose the skeptical loss that reduces the error in the estimation of correction parameters.
Finally, we propose the algorithm that updates the correction parameters during the training.

\subsection{Posterior Correction}
As we discussed in the previous section, the first moment of the gradient on a dataset is the expectation of that in the distribution given by the dataset's corresponding generation model.
The model \(G_{\tilde{D}}\) is available to compute the expectation of gradient in practice, because we can simulate it by computing the average of the gradient on the dataset or a mini-batch of the dataset.
However, the right expectation of the gradient should be computed in the distribution given by \(G_D\).
Therefore, we first figure out how the right expectation could be computed in the wrong distribution.
Notice that we optimize the model by maximizing its likelihood.

\begin{align}
	\begin{split}
		\label{equ:posterior0}
		& \mathbb{E}_{(\x, y) \sim p(G_D(\x, y))} \nabla_{\param} \log(p(y | \x; \param)) \\
		= & \sum\limits_{(\x, y) \in \mathcal{X} \times \mathcal{Y}} p(G_D(\x, y)) \nabla_{\param} \log(p(y | \x; \param)) \\
		= & \sum\limits_{(\x, y) \in \mathcal{X} \times \mathcal{Y}} p(G_D(\x, y)|G_D(\x)) p(G_D(\x)) \nabla_{\param} \log(p(y | \x; \param)) \\
		= & \sum\limits_{(\x, y) \in \mathcal{X} \times \mathcal{Y}} p(G_D(\x, y)|G_D(\x))p(G_{\tilde{D}}(\x)) \nabla_{\param} \log(p(y | \x; \param))
	\end{split}
\end{align}

Because \(G_D(\x) \equiv G_{\tilde{D}}(\x)\), we can derive the equation Eq. (\ref{equ:transform}). 
Then, we substitute it into the Eq. (\ref{equ:posterior0}), and we can get one of the two important forms of distribution correction Eq. (\ref{equ:posterior}).
Here, we simplify \(p(G_D(\x, y) | G_{\tilde{D}}(\x, \ty))\) as \(post(\x, y, \ty)\)

\begin{align}
	\small
	\begin{split}
	\label{equ:transform}
	& p(G_D(\x, y) | G_D(\x)) = p(G_D(\x, y)) p(G_D(\x))^{-1} \\
	= & \sum\limits_{\ty \in \mathcal{Y}} p(G_D(\x, y) | G_{\tilde{D}}(\x, y)) p(G_{\tilde{D}}(\x, \ty)) p(G_{\tilde{D}}(\x))^{-1} \\
	= & \sum\limits_{\ty \in \mathcal{Y}} p(G_D(\x, y) | G_{\tilde{D}}(\x, \ty)) p(G_{\tilde{D}}(\x, \ty) | G_{\tilde{D}}(\x))
	\end{split}
\end{align}

\begin{align}
	\small
	\begin{split}
	\label{equ:posterior}
	& \mathbb{E}_{(\x, y) \sim p(G_D(\x, y))} \nabla_{\param} \log(p(y | \x; \param)) \\
	= & \sum\limits_{(\x, y, \ty) \in \mathcal{X} \times \mathcal{Y} \times \mathcal{Y}} post(\x, y, \ty) p(G_{\tilde{D}}(\x, \ty)) \nabla_{\param} \log(p(y | \x; \param)) \\
	= & \mathbb{E}_{(\x, \ty) \sim p(G_D(\x, \ty))} \sum\limits_{y \in \mathcal{Y}} post(\x, y, \ty) \nabla_{\param} \log(p(y | \x; \param))
	\end{split}
\end{align}

We call the form of Eq. (\ref{equ:posterior}) posterior correction since it requires the posterior probabilities \(p(G_D(\x, y) | G_{\tilde{D}}(\x, \ty))\) for the correction.
There are many useful methods to estimate \(post(\x, y, \ty)\), and the quality of the estimation is very important.
For example, we can relax the posterior probabilities to \(p(G_D(y)|G_{\tilde{D}}(\ty))\), which is less difficult to estimate.
We have given a method of updating the estimation of the conditional probabilities \(\hat{p}(G_{\tilde{D}}(\ty)|G_D(y))\) during training later in this section, which can also estimate \(p(G_D(y)|G_{\tilde{D}}(\ty))\) with some modification.

\subsection{Conditional Correction}
Using the Bayes' theorem to unfold the posterior possibilities, we found another approach of computing \(post(\x, y, \ty)\), as shown in Eq. (\ref{equ:conditional0}).

\begin{align}
\begin{split}
	\label{equ:conditional0}
	& p(G_D(\x, y) | G_{\tilde{D}}(\x, \ty)) \\
	= & \frac{p(G_{\tilde{D}}(\x, \ty) | G_D(\x, y)) p(G_D(\x, y))}{\sum\limits_{y' \in \mathcal{Y}} p(G_{\tilde{D}}(\x, \ty) | G_D(\x, y')) p(G_D(\x, y'))} \\
	= & \frac{p(G_{\tilde{D}}(\x, \ty) | G_D(\x, y)) p(G_D(\x, y) | G_D(\x))}{\sum\limits_{y' \in \mathcal{Y}} p(G_{\tilde{D}}(\x, \ty) | G_D(\x, y')) p(G_D(\x, y') | G_D(\x))}
\end{split}
\end{align}

Substitute the Eq. (\ref{equ:conditional0}) into the posterior correction form Eq. (\ref{equ:posterior}).
We can then derive the other important form of distribution correction, which is named conditional correction Eq. (\ref{equ:conditional}).
Again, we simplify \(p(G_{\tilde{D}}(\x, \ty) | G_D(\x, y))\) as \(cond(\x, y, \ty)\).
Notice that \(\sum\limits_{y' \in \mathcal{Y}} cond(\x, y', \ty) p(G_D(\x, y'))\) is invariant when \((\x, \ty)\) is fixed.

\begin{align}
\begin{split}
	\label{equ:conditional}
	& \sum\limits_{y \in \mathcal{Y}} post(\x, y, \ty) \nabla_{\param} \log(p(y | \x; \param)) \\ 
	= & \frac{ \sum\limits_{y \in \mathcal{Y}} cond(\x, y, \ty) p(G_D(\x, y) | G_D(\x)) \nabla_{\param} \log(p(y | \x; \param))}{\sum\limits_{y' \in \mathcal{Y}} cond(\x, y', \ty) p(G_D(\x, y') | G_D(\x))}
\end{split}
\end{align}

In this form, we have two expressions two estimate: the conditional probabilities \(p(G_{\tilde{D}}(\x, \ty) | G_D(\x, y'))\), and the right label distribution \(p(G_D(\x, y) | G_D(\x))\).
Although there are more figures to estimate, we think it is the better form.
It's time to be skeptical.
If we use the model's present prediction to substitute every \(p(G_D(\x, y) | G_D(\x))\), we will have the following derivation.

\begin{align}
\begin{split}
\label{equ:forward0}
	& \frac{ \sum\limits_{y \in \mathcal{Y}} cond(\x, y, \ty) p(y | \x; \param) \nabla_{\param} \log(p(y | \x; \param))}{\sum\limits_{y' \in \mathcal{Y}} cond(\x, y', \ty) p(y' | \x; \param)} \\
	= & \frac{\sum\limits_{y \in \mathcal{Y}} cond(\x, y, \ty) \nabla_{\param} p(y | \x; \param)}{\sum\limits_{y' \in \mathcal{Y}} cond(\x, y', \ty) p(y' | \x; \param)} \\
	= & \nabla_{\param} \log \sum\limits_{y \in \mathcal{Y}} cond(\x, y, \ty) p(y | \x; \param)
\end{split}
\end{align}

\(cond(\x, y, \ty)\) can be relaxed into a simpler \(p(G_{\tilde{D}}(\ty) | G_D(y))\).
This implementation of conditional correction is equivalent to the forward loss correction proposed by \cite{lca17}.
Thus, distribution correction gives an explanation of the method.
Moreover, we can tell why this implementation performs beyond expectation.
The relaxation ignores the label noise cased by individual feature, but this error can be reduced by the multiplication of the model's prediction and the normalization in the denominator of the RHS of Eq. (\ref{equ:conditional}).

\subsection{Learning to be Skeptical}
Although the previous method is excellent, there is still room for improvement, especially for the quality of estimation.
The term \(cond(\x, y, \ty) p(G_D(\x, y) | G_D(\x))\) in the numerator of Eq. (\ref{equ:conditional}) add weights to each label, while the denominator normalize the sum of weights into a unit.
We still estimate the weights in the numerator by \(p(G_{\tilde{D}}(\ty)　| G_D(y)) p(y | \x; \param)\).
The error is magnified when \(\sum\limits_{y \in \mathcal{Y}} p(G_{\tilde{D}}(\ty)　| G_D(y)) p(y | \x; \param)\) is significantly below \(1\).
Actually, the sum of weights is \(p(G_{\tilde{D}}(\x, \ty) | p(G_{\tilde{D}}(\x))\), which is usually \(1\) or marginally below \(1\) in practice.
For this reason, we can estimate the denominator by \(1\) or a normally trained model's prediction.
However, those methods have a great influence to the distribution, so we want to find a way do it mildly.
Therefore, we decide to estimate the denominator by a scaled sum of \(p(G_{\tilde{D}}(\ty)　| G_D(y)) p(y | \x; \param)\).

Suppose the magnification function \(f\) takes two inputs: \(p = \sum\limits_{y \in \mathcal{Y}} p(G_{\tilde{D}}(\ty)　| G_D(y)) p(y | \x; \param)\) and \(k = G_{\tilde{D}}(y)\).
If there are the same number of instances for each label in the noisy dataset, \(k = |\mathcal{Y}^{-1}|\).
The scaling function has some restrictions. 
First, it magnifies a probability and its result is also a probability (Eq. \ref{equ:restriction1}). Second, the function should be monotone increasing, and the result should not be less than \(p\) (Eq. \ref{equ:restriction2}).
The last but not least, the scale rate on each label should be the same when \(p=k\) (Eq. \ref{equ:restriction3}).
The last rule ensures that the magnification is not mainly affected by \(G_{\tilde{D}}(y)\).

\begin{align}
	\begin{split}
		\label{equ:restriction1}
		f(0; k) = 0, f(1; k) & = 1; f(p; k) \in [0, 1]
	\end{split}
	\\
	\begin{split}
	\label{equ:restriction2}
		p_1 < p_2 \Rightarrow f(p_1; k) & < f(p_2; k); f(p; k) \ge p 
	\end{split}
	\\
	\begin{split}
	\label{equ:restriction3}
		k_2f(k_1; k_1) & = k_1f(k_2; k_2)
	\end{split}
\end{align}

We found a solution Eq. (\ref{equ:skgrad}) that meets those restrictions above, where \(\beta\) is a hyper-parameter .
The magnification is more significant when \(\beta\) is lower.
Just like we did in Eq. (\ref{equ:forward0}), we can also derive the corresponding loss function Eq. (\ref{equ:skloss}).
This loss function has a wonderful attribute that \(\ln p - L_{\mathcal{SK}}(p) = O((p-1)^2)\), which means that it is similar to log loss when \(p \to 1\).
It is a bounded function that does not go to infinity when \(p\to 0\).
We name it skeptical loss, because it reduces the proportion of suspicious data samples in calculating the expectation of the gradient.
Those suspicious data has significantly low  \(p(G_{\tilde{D}}(\ty)　| G_D(y)) p(y | \x; \param)\), which is derived from the model's prediction.
Therefore, the model is skeptical again when we use the skeptical loss.

\begin{align}
\begin{split}
\label{equ:skgrad}
f(p;k) &= \frac{p^{1-k^\beta}}{1-k^\beta\ln p}, \beta \in (0, 1)
\end{split}
\\
\begin{split}
\label{equ:skloss}
L_{\mathcal{SK}}(p;k) &=p^{k^\beta}(2k^{-\beta}-\ln p), \beta \in (0, 1)
\end{split}
\end{align}

\begin{figure}
	\centering
	\includegraphics[width=0.46\textwidth]{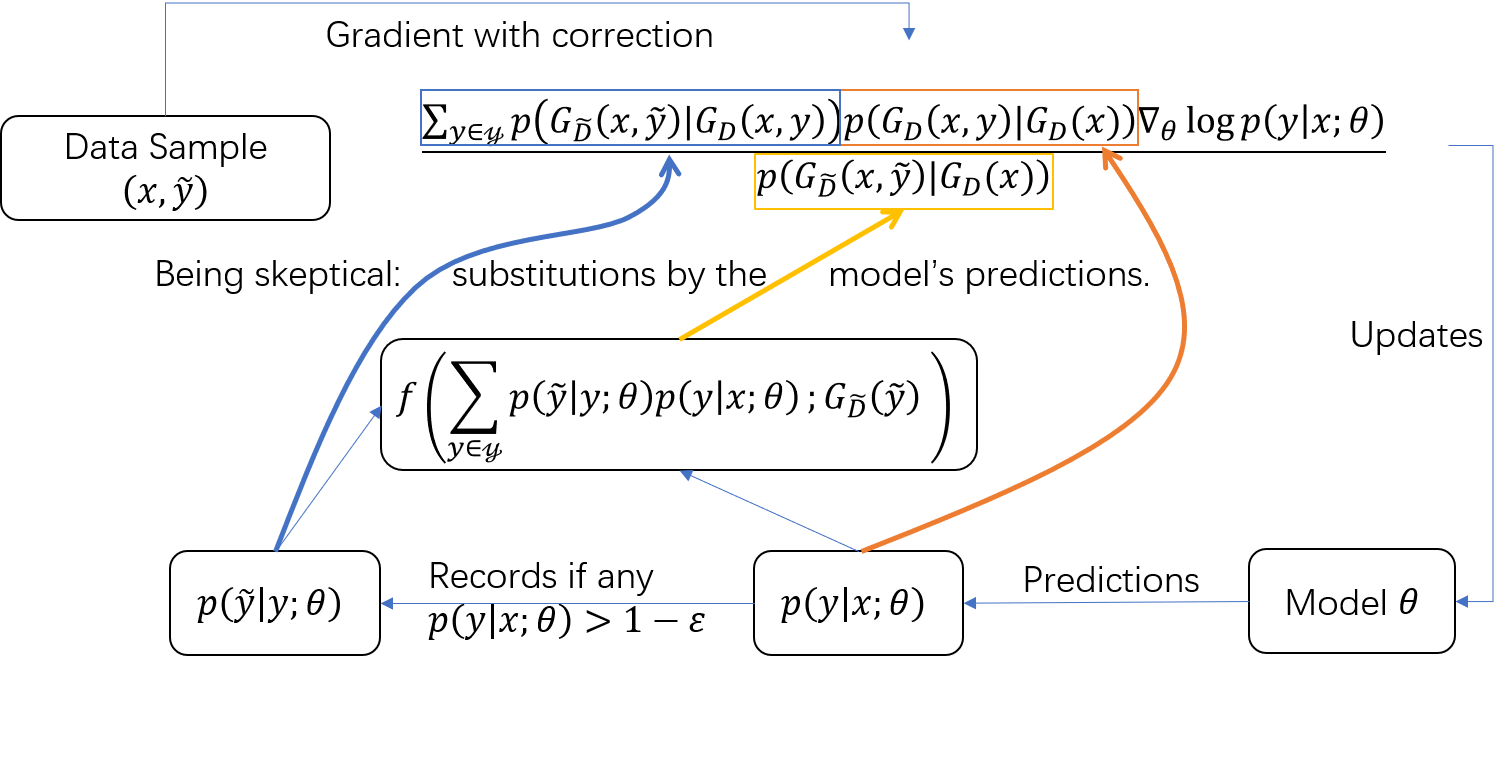}
    \caption{The three substitutions of being skeptical.}
    \label{fig:skeptial}
\end{figure}

Finally, we propose a practical algorithm of estimating \(p(G_{\tilde{D}}(\ty)|G_D(y))\).
Here, we still assume that the model's prediction during training is the right prediction.
Because the prediction of the model is changing during the training process, the value of the estimation \(\hat{p}(G_{\tilde{D}}(\ty)|G_D(y))\) is also invariant for the safety reason.
We simplify the notation \(\hat{p}(G_{\tilde{D}}(\ty)|G_D(y))\) as \(\hat{\mathcal{T}}_{\ty, y}\).
Those parameters are initialized by \(\mathbb{I}_{\ty}(y)\), which means that we treat the noisy dataset as the right one in the beginning. 
Then, if the model has a prediction of \(p(y | \x) > 1 - \epsilon\), we think the model is confident to this prediction.
The estimation \(\hat{\mathcal{T}}_{\ty, y}\) should be updated no matter whether there is a conflict between a confident prediction and the label in the noisy dataset.
We update the parameters by \(\hat{\mathcal{T}}_{y', y}^{t+1} = \gamma\hat{\mathcal{T}}_{y', y}^{t} + (1 - \gamma)\mathbb{I}_{\ty}(y')\) for each \(y' \in \mathcal{Y}\), which ensures \(\sum\limits_{\ty \in \mathcal{Y}} \hat{\mathcal{T}}_{\tilde{y}, y} = 1\).
The greater \(\gamma\) and smaller \(\epsilon\) are safer, and we recommend the safer parameters.
The three substitutions are shown in Fig.(\ref{fig:skeptial}). Alg.(\ref{alg:learning}) shows the overall algorithm of learning with conditional correction using skeptical loss.

\begin{algorithm}
	\caption{Learning with Conditional Correction Using Skeptical Loss}
	\label{alg:learning}
	\begin{algorithmic}[1]
		\STATE{Initialize \(\param\); \(\hat{\mathcal{T}}_{\ty, y} \gets \mathbb{I}_{\ty}(y)\) for each \((\ty, y)\in\mathcal{Y}\times\mathcal{Y}\)}
		\FOR{each minibatch \(B\subseteq\tilde{D}\)}
		\STATE{\(\mathrm{grad} \gets 
			\sum\limits_{(\x, \ty)\in B} \nabla_{\param} L_{\mathcal{SK}}(\sum\limits_{y\in\mathcal{Y}}\hat{\mathcal{T}}_{\ty, y}p(\ty|\x; \param); |\mathcal{Y}|^{-1})
			\)}
		\STATE{\(\param\) = update(\(\param\), \(\mathrm{grad}\))}
		\FOR{each \((\x, \ty)\in B\)}
		\STATE{\(y\gets\mathop{\mathrm{argmax}}\limits_{y}p(y | \x; \param)\)}
		\IF{\(p(y | \x; \param) > 1-\epsilon\)}
		\STATE{\(\hat{\mathcal{T}}_{y', y}^{t+1} = \gamma\hat{\mathcal{T}}_{y', y}^{t} + (1 - \gamma)\mathbb{I}_{\ty}(y')\) for each \(y'\in\mathcal{Y}\)}
		\ENDIF
		\ENDFOR
		\ENDFOR
	\end{algorithmic}
\end{algorithm}

\section{Experiments}
In this section, we first introduce our method of generating noisy datasets.
Second, we give details about the experiment settings.
We have done experiments on both our elaborately generated noise and symmetric noise.
Finally, we show the empirical result and give analysis.

\subsection{Datasets Preparation}
Generating the noisy datasets from existed ones is a useful way to study label noise.
We know the true labels of the generated datasets, which are reliable feedback. 
The real-life label noise consists of several types of the noise model. 
Among them, it is difficult to generate the model of noise that is confusing, that the wrong label is the most likely one among the wrong labels. 
For example, the wrong label of a walking cat picture is more likely to be misclassified into a standing lion than a laying hen for the confusing label noise, but these probabilities are equal for the symmetric model. 
However, it seems impossible to manually generate these labels by human force. 
Thus, we propose the following algorithm to generate noisy datasets.

Given the original dataset \(D\) and its subset \(B\subset D\) of \(|B|=p_e|D|\), the noisy dataset generated is \(\tilde{D} = \tilde{B} \cup (D \setminus B)\), where \(\tilde{B} = \mathop{\mathrm{argmax}}\limits_{B' \in \mathcal{B}} p(B' | G_{\tilde{D}})), \mathcal{B}=\{B' | \forall (\x, y)\in B, \exists\tilde{y}\neq y, (\x, \tilde{y})\in B' \mathrm{\ and\ } |B|=|B'| \}\).
In practice, we can first generate the wrong dataset and then select labels from these two datasets according to \(p_e\).
The wrong labels are selected as the first or second top labels predicted by the model trained by dataset \(\tilde{D}\) without any correction.
The following algorithm (\ref{alg:datagen}) describes the process above.
Notice that the models trained here should have the same network structure as the target models.
This generated label noise should be confusing to the target models.
Therefore, we call it confusing label noise.

\begin{algorithm}
	\caption{Generate Dataset with Confusing Label Noise}
	\label{alg:datagen}
	\begin{algorithmic}[1]
		\STATE{\(\bar{D}\gets\varnothing; \tilde{D}\gets \varnothing\)}
		\STATE{\(\hat{\param}_D \gets \mathop{\mathrm{argmax}}\limits_{\param} \mathcal{L}(\param | D)\)}
		\FOR{each \((\x, y)\in{D}\)}
		\STATE{\(y_1, y_2 \gets \mathop{\mathrm{argmax_2}}
			\limits_{y' \in \mathcal{Y}} p(y' | \x; \hat{\param}_D)\)}
		\STATE{\(\bar{D}\gets\) \textbf{if} \(y_1=y\) \textbf{then} \(\bar{D}\cup\{(\x, y_2)\}\) \textbf{else} \(\bar{D}\cup\{(\x, y)\}\)}
		\ENDFOR
		\STATE{Randomly sample \(D'\subset D\), where \(|D'|=p_e|D|\)}
		\FOR{each \((\x, y)\in D\) and corresponding \((\x, \ty) \in \bar{D}\)}
		\STATE{\(\tilde{D}\gets\) \textbf{if} \((\x, y) \in D'\) \textbf{then} \(\tilde{D} \cup \{\x, \ty\}\) \textbf{else} \(\tilde{D} \cup\{(\x, y)\}\)}
		\ENDFOR
	\end{algorithmic}
\end{algorithm}

\begin{figure*}
	\centering
	\begin{tabular}{cc}
		\includegraphics[width=.40\textwidth]{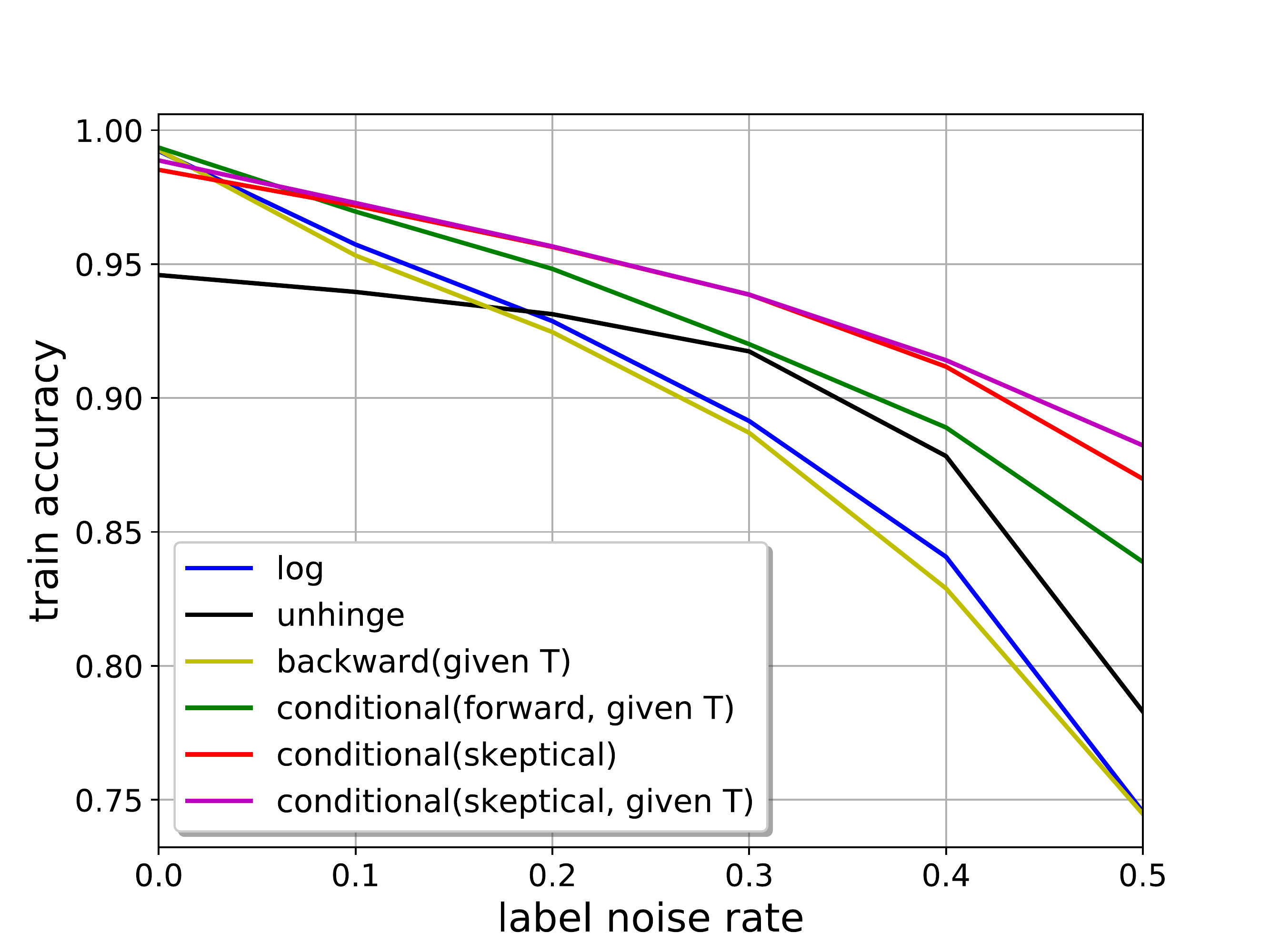} & \includegraphics[width=.40\textwidth]{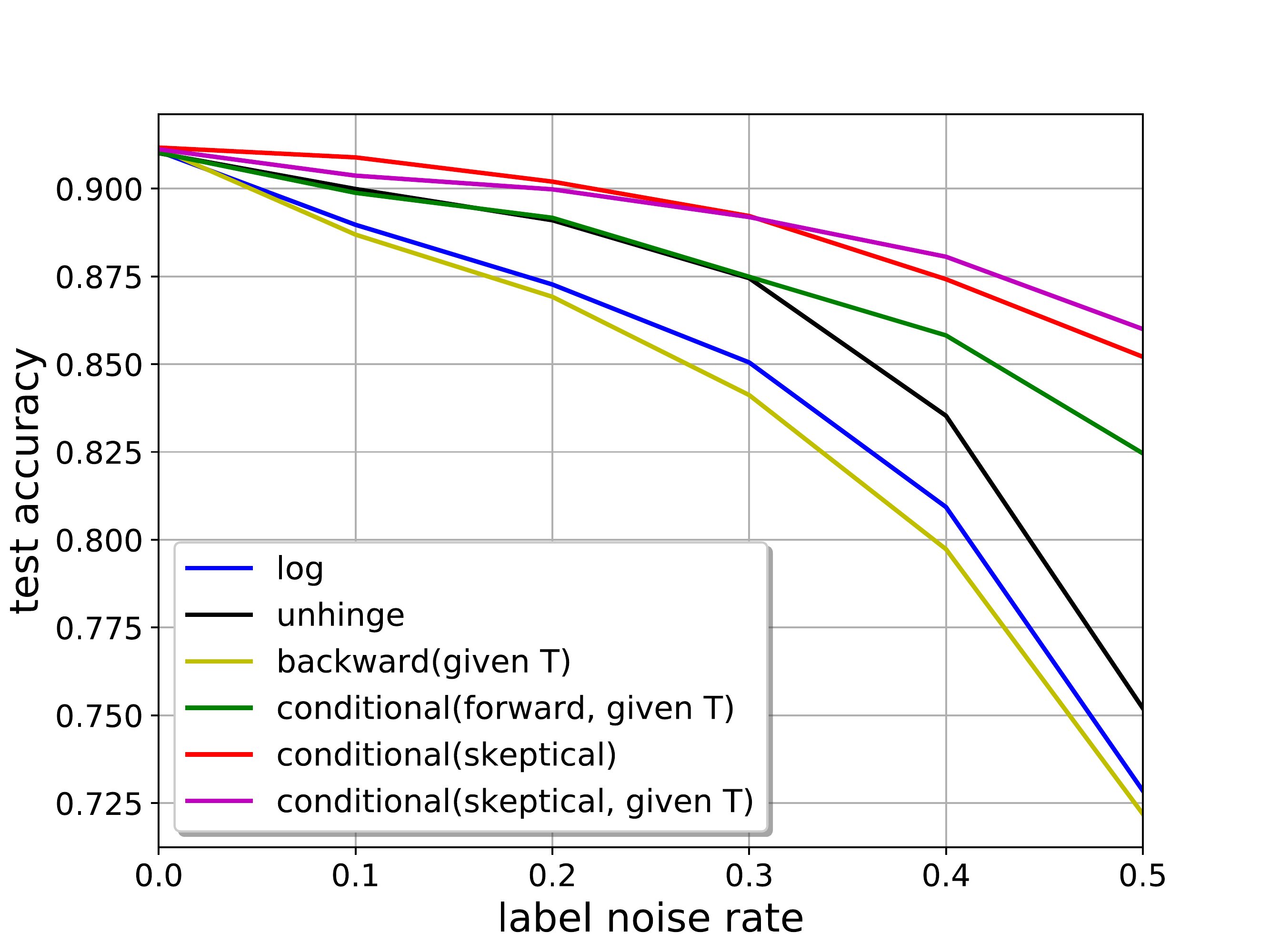} \\
		FMNIST(S) Train Acc & FMNIST(S) Test Acc \\
		\includegraphics[width=.40\textwidth]{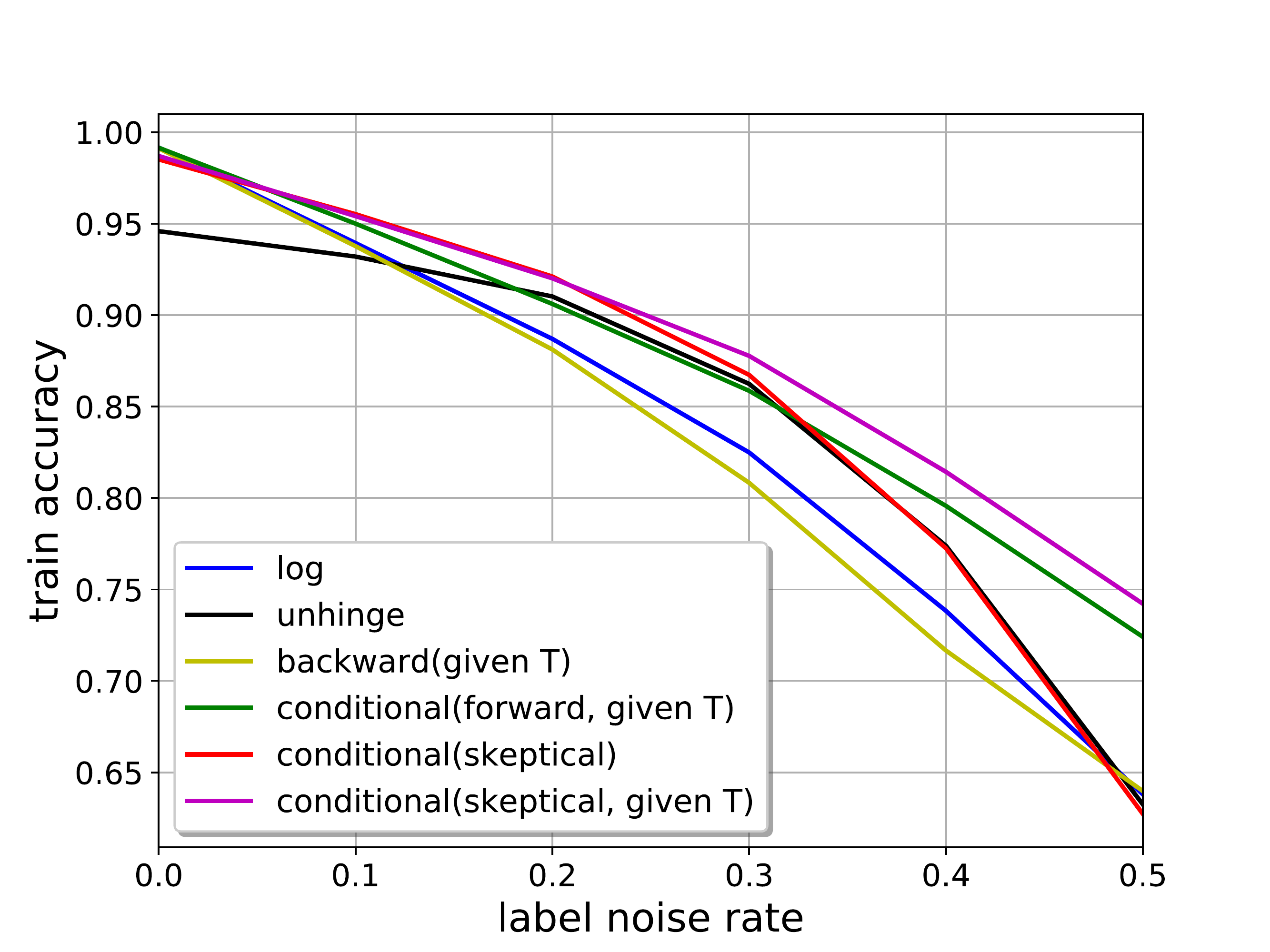} &
		\includegraphics[width=.40\textwidth]{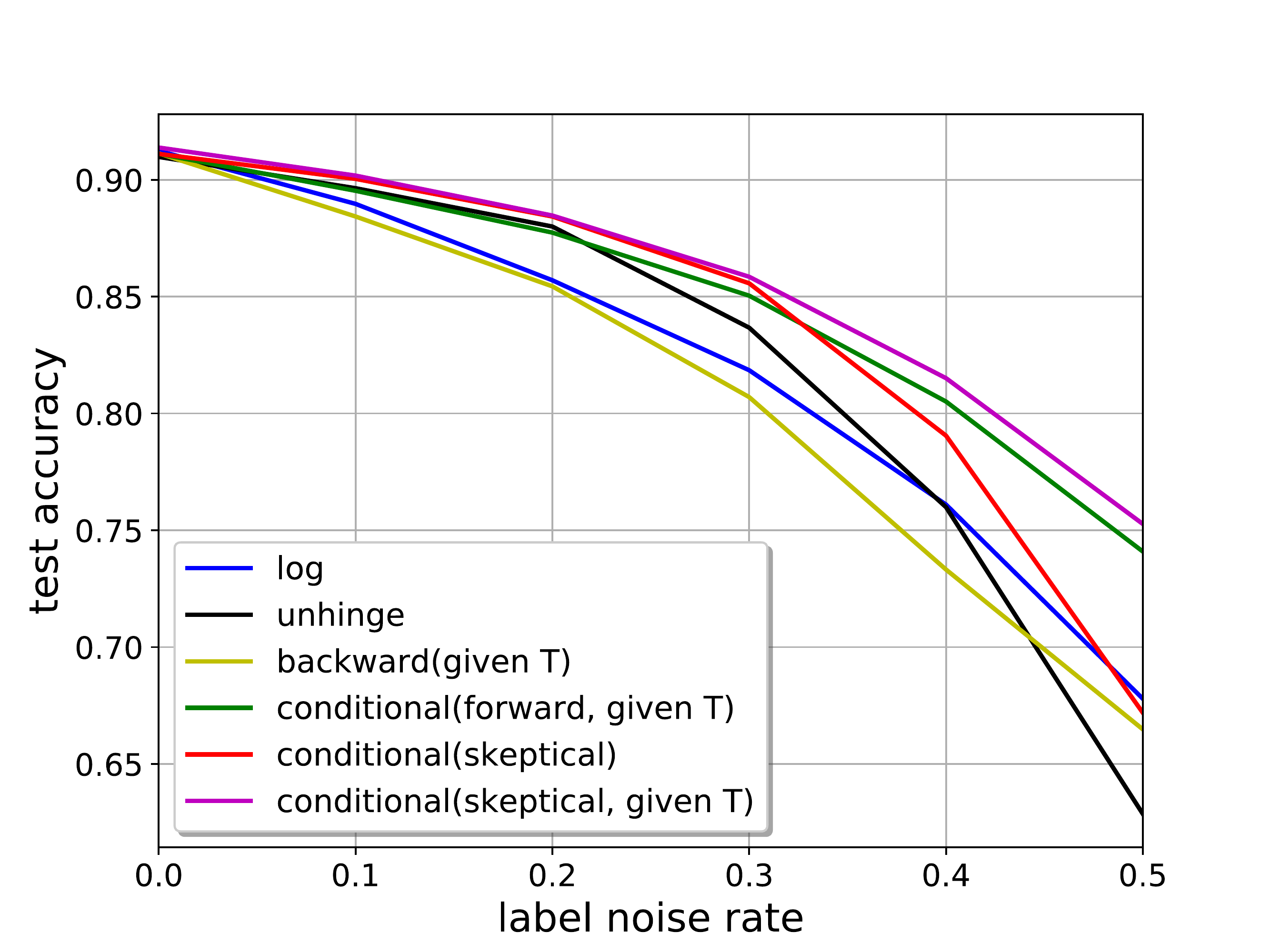} \\ 
		FMNIST(C) Train Acc & FMNIST(C) Test Acc \\
		\includegraphics[width=.40\textwidth]{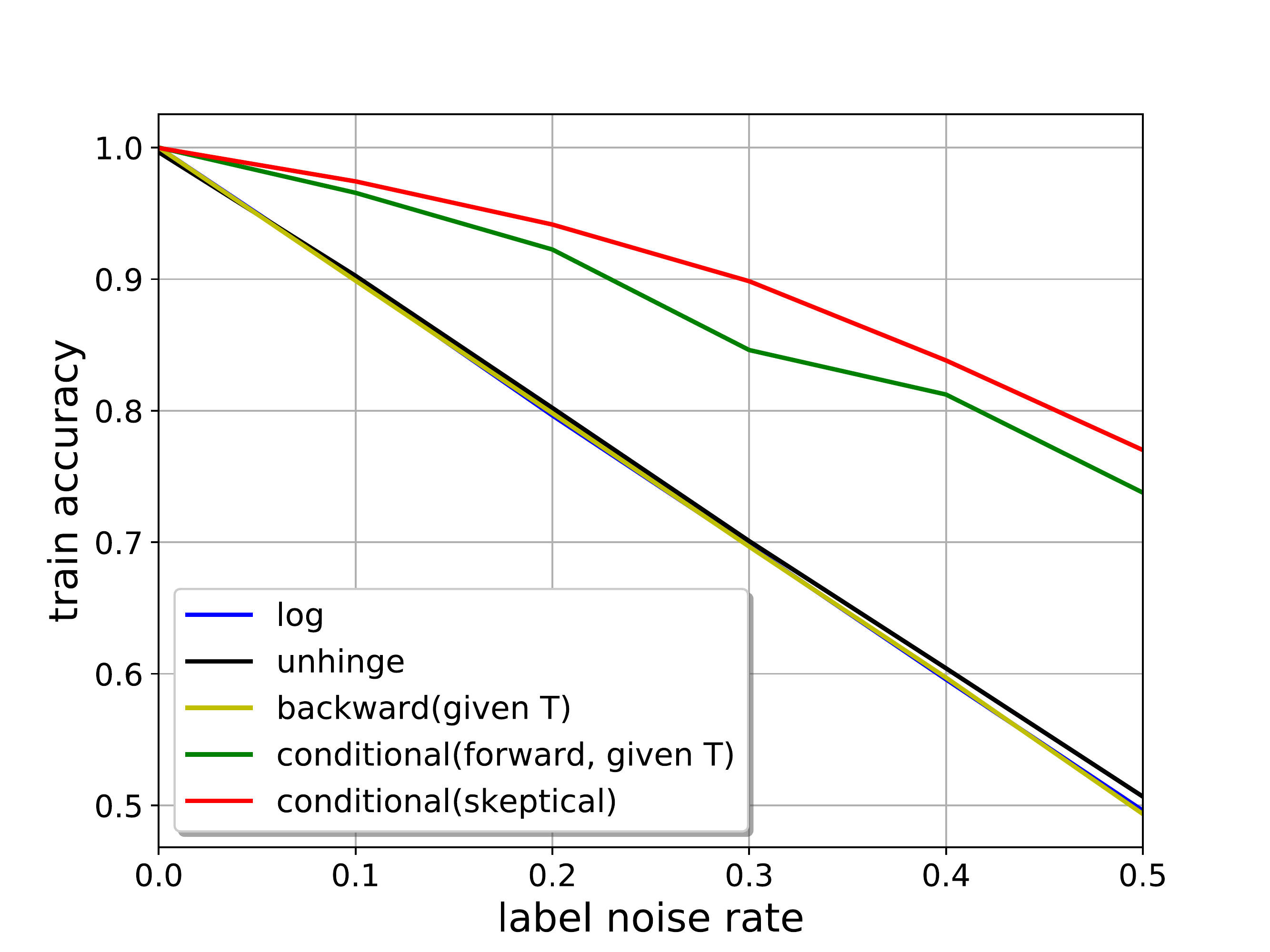} & \includegraphics[width=.40\textwidth]{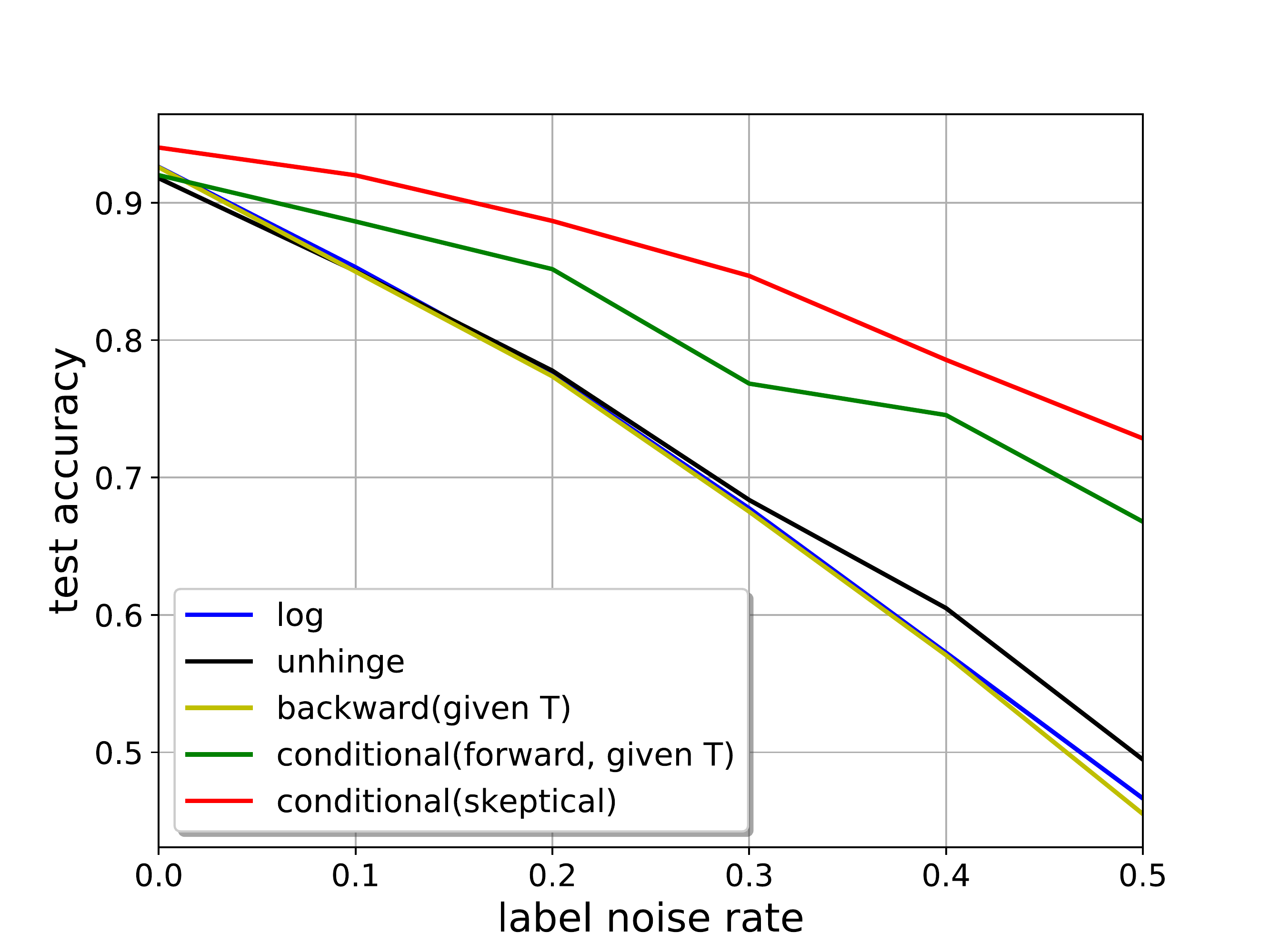} \\
		CIFAR-10(S) Train Acc & CIFAR-10(S) Test Acc \\
		\includegraphics[width=.40\textwidth]{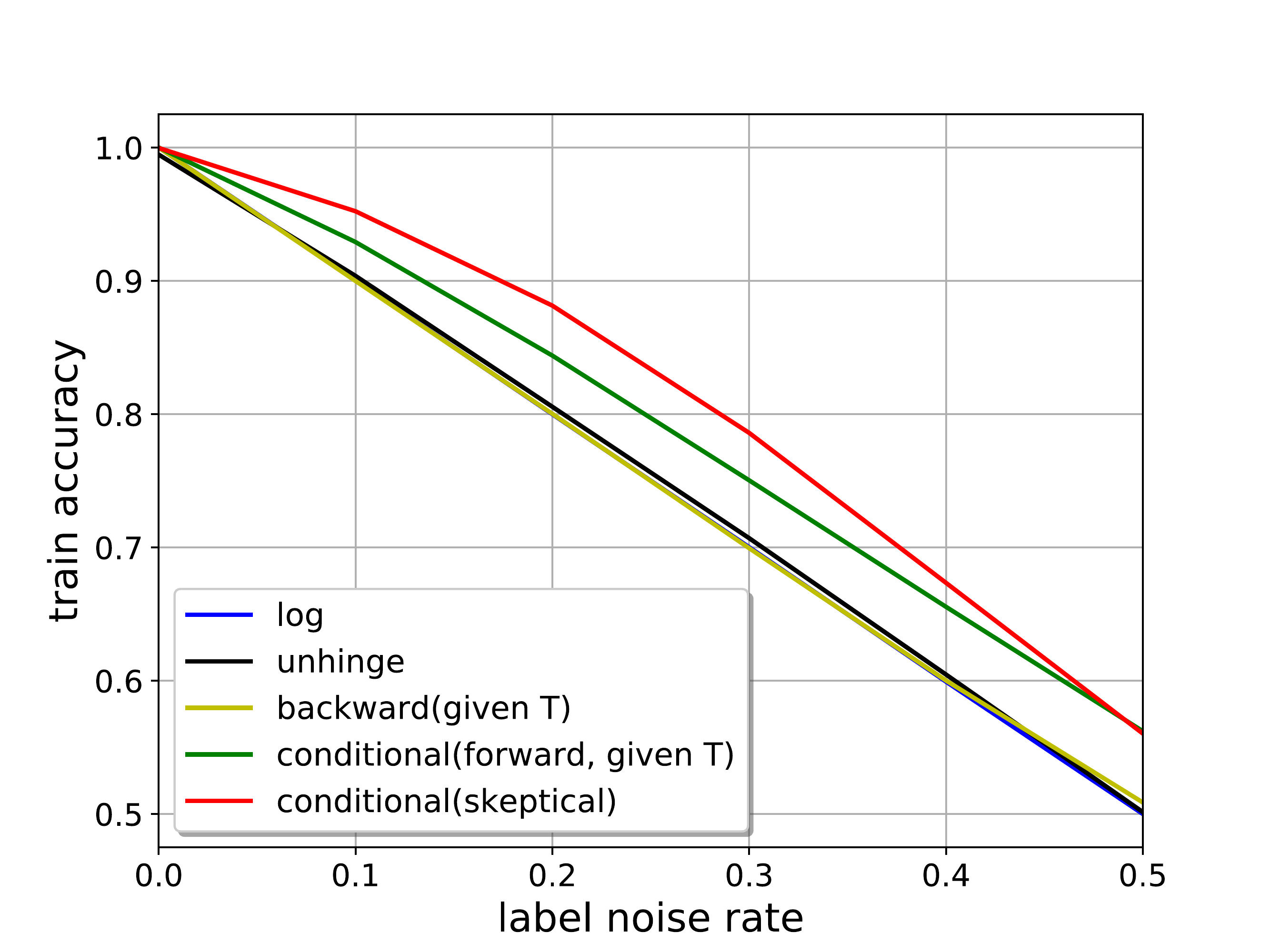} &
		\includegraphics[width=.40\textwidth]{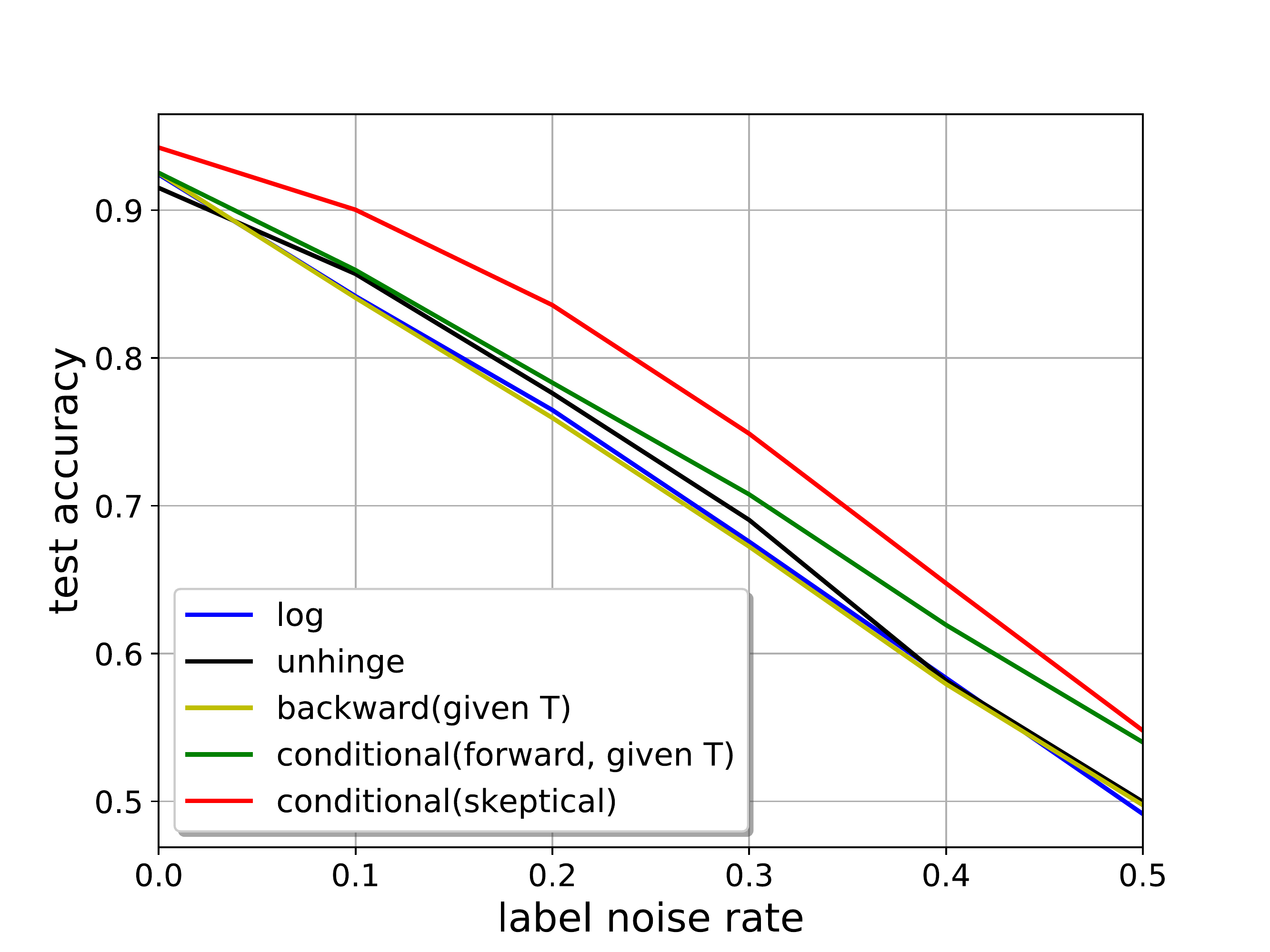} \\ CIFAR-10(C) Train Acc & CIFAR-10(C) Test Acc \\
	\end{tabular}
	\caption{Performance with increasing label noise ratio. "C" stands for "Confusing" and "S" stands for "Symmetric".}
	\label{fig:figures}
\end{figure*}

\begin{table*}
	\caption{Detailed comparison under several certain settings. All figures are the partial mean in 5 runs.}
	\label{tab:detail}
	\centering
	\begin{tabular}{lllllll}
		\toprule
		Method & Dataset & Noise & Test Err & Train Err & Precision & Recall \\ 
		\midrule
		Log & FMNIST & C0.4 & 0.2390 & 0.2617 & 0.8179 & 0.4206 \\
		Unhinged & FMNIST & C0.4 & 0.2401 & 0.2261 & 0.7561 & 0.5907 \\
		Backward (\(\mathcal{T}\)) & FMNIST & C0.4 & 0.2668 & 0.2834 & 0.6755 & 0.5080 \\
		Conditional (Forward, \(\mathcal{T}\)) & FMNIST & C0.4 & 0.1949 & 0.2043 & 0.8225 & 0.5932 \\
		Conditional (Skeptical, \(\mathcal{T}\)) & FMNIST & C0.4 & \textbf{0.1854} & \textbf{0.1858} & \textbf{0.8451} & \textbf{0.6270} \\
		Conditional (Skeptical, \(\mathcal{\hat{T}}\)) & FMNIST & C0.4 & 0.2075 & 0.2275 & 0.8707 & 0.4868 \\
		\midrule
		Log & CIFAR-10 & C0.4 & 0.4165 & 0.4007 & 0.3686 & 0.0022 \\
		Unhinged & CIFAR-10 & C0.4 & 0.4183 & 0.3954 & 0.3723 & 0.0436 \\
		Backward (\(\mathcal{T}\)) & CIFAR-10 & C0.4 & 0.4206 & 0.3999 & 0.3749 & 0.0042 \\
		Conditional (Forward, \(\mathcal{T}\)) & CIFAR-10 & C0.4 & 0.3806 & 0.3446 & 0.6993 & 0.1642 \\
		Conditional (Skeptical, \(\mathcal{\hat{T}}\)) & CIFAR-10 & C0.4 & \textbf{0.3524} & \textbf{0.3266} & \textbf{0.7045} & \textbf{0.2289} \\
		\midrule
		Log & ILSVRC & S0.2 & 0.5898 & 0.4841 & 0.1522 & 0.4384 \\
		Backward (\(\mathcal{T}\)) & ILSVRC & S0.2 & 0.5827 & 0.4859 & 0.1540 & 0.4383 \\
		Conditional (Forward, \(\mathcal{T}\)) & ILSVRC & S0.2 & 0.5396 & \textbf{0.3389} & \textbf{0.2246} & \textbf{0.4890} \\
		Conditional (Skeptical, \(\mathcal{T}\)) & ILSVRC & S0.2 & \textbf{0.5384} & 0.3491 & 0.2177 & 0.4860 \\
		Conditional (Skeptical, \(\mathcal{\hat{T}}\)) & ILSVRC & S0.2 & 0.5423 & 0.3954 & 0.1920 & 0.4707 \\
		\bottomrule
	\end{tabular}
\end{table*}

We have done tests on several image classification tasks in order to do baseline comparisons, and we will show the typical ones following in this section. 
Before we did the tests, we use MNIST \cite{lec98}, CIFAR-100 \cite{cif09}, and ILSVRC \cite{img12} datasets to tune the hyper-parameters of model architecture, training process, and our method.
With the same settings, we tested these models on FMNIST \cite{fmn17}, CIFAR-10 \cite{cif09} and downsampled ILSVRC \cite{dsp17}.
To show how the methods are adaptive to various types of label noise, we generate datasets of both symmetric and confusing noise.
For each type of noise, there are datasets of different \(p_e\).
Moreover, We prepared 5 different generated datasets for each type of noise.
All the figures are calculated as the partial mean of the five results (the mean without the highest and lowest one).
We design the experiments to simulate the situation that no any information except the noisy training dataset is given.

\subsection{Empirical Tests}
In the first example, we use CIFAR-10 as dataset and models are trained by WRN-28-2 \cite{wrn16} with \(3\times3\) convolutions.
WRNs are regularized architecture of residual networks \cite{res15}\cite{res16}.
We optimize the networks by gradient descent with \(0.9\) momentum, \(0.001\) scale of L2-regularization \cite{l2r09}, and batch size \(128\)  in 200 epochs.
Learning rate starts with \(0.1\) and decays by \(0.2\) at the number 60, 120, and 160 epochs.
We choose these hyperparameters because we think they are widely accepted and open.

We also use FMNIST, which is designed as an alternative to MNIST, as an example of the simple dataset.
Using widely accepted architecture again, we select LeNet-5 \cite{lec98} as the model.
Although our implementation of distribution correction does not consider the second moment of first gradient, we still try Adam \cite{ada14} with \(\mathrm{lr}=10^{-3}, \beta_1=0.9, \beta_2=0.999\) to optimize the model.
We extend the training process to 200 epochs, and we found the performance of log loss then is much worse than 50 epochs.
Therefore, our tests of FMNIST are done in extreme condition, which magnifies the difference.
In application, one can actually adopt early-stop \cite{ear17} (without validation set) or annealing \cite{ann94} methods to find the best number of iterations.

We train the WRN-37-2 networks on downsampled \(64\times64\) ILSVRC for the last case.
We use this downsampled variant of ILSVRC because we find it very sensitive to the label noise.
The networks are optimized similar to the previous WRN-28-2, with \(0.0005\) scale of L2-regularization in 50 epochs. 
Learning rate starts with \(0.02\) and decays by \(0.2\) at the number 20, 35, and 45 epochs.

For conditional correction with skeptical loss, we use \(\beta = 0.2, \gamma=0.9999, \epsilon=0.1\) for all the tests.
We have found out the best hyper-parameters for each case (MNIST, CIFAR-10, and ILSVRC).
It is important that we pick the safest one for each parameter: the largest \(\beta\) and \(\gamma\), and the smallest \(\epsilon\) among the best ones.
We also test our method with the true \(\mathcal{T}\), and the same true \(\mathcal{T}\) is given to all the baselines.
Other methods are tested under the same settings. 
We have implemented unhinged loss \cite{unh15}, as well as the backward and forward correction as special cases of distribution correction. 
Because of its open bound, we add batch normalization \cite{bat15} and great L2-regularization to make unhinge loss work for multiple classifications.

\subsection{Result and Analysis}

The figures in fig. (\ref{fig:figures}) show the power of label noise resistance with the increasing \(p_e\). 
The table (\ref{tab:detail}) gives detailed information for some certain conditions.
These results meet our expectation that our method is adaptive to various types of label noise, which has verified our assumptions for distribution correction.
Besides, we found that the log loss has nearly no resistance to confusing label noise, but there is still a little resistance for symmetric label noise.
This experiment result proves that the confusing datasets are really difficult for maximum-likelihood training.

Although the performance of \(\mathcal{T}\) estimation is not perfect for downsampled ILSVRC, we are still conservative about the update ratio. 
It implies that \(\hat{\mathcal{T}}\) has not reached its destination before the training process stopped.
Besides, we found the noise accuracy increases slower at the beginning when trained with the skeptical loss, which means that we should give enough iterations before the learning rate decay in residual networks. 
For the same reason, we do not recommend to use the skeptical loss alone without any correction.
\section{Conclusion}
In this paper, we proposed the distribution correction approach to make models robust to label noise. 
The empirical result showed how the simple and online implementation of conditional correction with skeptical loss is effective. 
This approach could help people to analyze learning with label noise without any prior information given.
Also, we give a solution to generate feature-dependent label noise, which can contribute to further works on label noise.

\bibliographystyle{aaai.bst}
\bibliography{ref.bib}

\begin{thebibliography}{}

\bibitem[\protect\citeauthoryear{Angluin and Laird}{1988}]{lfn88}
Angluin, D., and Laird, P.
\newblock 1988.
\newblock Learning from noisy examples.
\newblock {\em Machine Learning} 2(4):343--370.

\bibitem[\protect\citeauthoryear{Chrabaszcz, Loshchilov, and
  Hutter}{2017}]{dsp17}
Chrabaszcz, P.; Loshchilov, I.; and Hutter, F.
\newblock 2017.
\newblock A downsampled variant of imagenet as an alternative to the cifar
  datasets.
\newblock {\em arXiv preprint arXiv:1707.08819}.

\bibitem[\protect\citeauthoryear{Cortes, Mohri, and Rostamizadeh}{2009}]{l2r09}
Cortes, C.; Mohri, M.; and Rostamizadeh, A.
\newblock 2009.
\newblock L2 regularization for learning kernels.
\newblock In {\em Proceedings of the Twenty-Fifth Conference on Uncertainty in
  Artificial Intelligence},  109--116.
\newblock AUAI Press.

\bibitem[\protect\citeauthoryear{Cristianini and Shawe-Taylor}{2000}]{bou00}
Cristianini, N., and Shawe-Taylor, J.
\newblock 2000.
\newblock {\em An introduction to support vector machines and other
  kernel-based learning methods}.
\newblock Cambridge university press.

\bibitem[\protect\citeauthoryear{Fr{\'e}nay and Verleysen}{2014}]{sur14}
Fr{\'e}nay, B., and Verleysen, M.
\newblock 2014.
\newblock Classification in the presence of label noise: a survey.
\newblock {\em IEEE transactions on neural networks and learning systems}
  25(5):845--869.

\bibitem[\protect\citeauthoryear{Ghosh, Kumar, and Sastry}{2017}]{rob17}
Ghosh, A.; Kumar, H.; and Sastry, P.
\newblock 2017.
\newblock Robust loss functions under label noise for deep neural networks.
\newblock In {\em AAAI},  1919--1925.

\bibitem[\protect\citeauthoryear{Goldberger and Ben-Reuven}{2016}]{nal16}
Goldberger, J., and Ben-Reuven, E.
\newblock 2016.
\newblock Training deep neural-networks using a noise adaptation layer.

\bibitem[\protect\citeauthoryear{He \bgroup et al\mbox.\egroup }{2015}]{res15}
He, K.; Zhang, X.; Ren, S.; and Sun, J.
\newblock 2015.
\newblock Delving deep into rectifiers: Surpassing human-level performance on
  imagenet classification.
\newblock In {\em Proceedings of the IEEE international conference on computer
  vision},  1026--1034.

\bibitem[\protect\citeauthoryear{He \bgroup et al\mbox.\egroup }{2016}]{res16}
He, K.; Zhang, X.; Ren, S.; and Sun, J.
\newblock 2016.
\newblock Deep residual learning for image recognition.
\newblock In {\em Proceedings of the IEEE conference on computer vision and
  pattern recognition},  770--778.

\bibitem[\protect\citeauthoryear{Ioffe and Szegedy}{2015}]{bat15}
Ioffe, S., and Szegedy, C.
\newblock 2015.
\newblock Batch normalization: Accelerating deep network training by reducing
  internal covariate shift.
\newblock {\em arXiv preprint arXiv:1502.03167}.

\bibitem[\protect\citeauthoryear{Kingma and Ba}{2014}]{ada14}
Kingma, D.~P., and Ba, J.
\newblock 2014.
\newblock Adam: A method for stochastic optimization.
\newblock {\em arXiv preprint arXiv:1412.6980}.

\bibitem[\protect\citeauthoryear{Krause \bgroup et al\mbox.\egroup
  }{2016}]{unr16}
Krause, J.; Sapp, B.; Howard, A.; Zhou, H.; Toshev, A.; Duerig, T.; Philbin,
  J.; and Fei-Fei, L.
\newblock 2016.
\newblock The unreasonable effectiveness of noisy data for fine-grained
  recognition.
\newblock In {\em European Conference on Computer Vision},  301--320.
\newblock Springer.

\bibitem[\protect\citeauthoryear{Krizhevsky and Hinton}{2009}]{cif09}
Krizhevsky, A., and Hinton, G.
\newblock 2009.
\newblock Learning multiple layers of features from tiny images.

\bibitem[\protect\citeauthoryear{Krizhevsky, Sutskever, and
  Hinton}{2012}]{img12}
Krizhevsky, A.; Sutskever, I.; and Hinton, G.~E.
\newblock 2012.
\newblock Imagenet classification with deep convolutional neural networks.
\newblock In {\em Advances in neural information processing systems},
  1097--1105.

\bibitem[\protect\citeauthoryear{Laine and Aila}{2016}]{pim16}
Laine, S., and Aila, T.
\newblock 2016.
\newblock Temporal ensembling for semi-supervised learning.
\newblock {\em arXiv preprint arXiv:1610.02242}.

\bibitem[\protect\citeauthoryear{LeCun \bgroup et al\mbox.\egroup
  }{1998}]{lec98}
LeCun, Y.; Bottou, L.; Bengio, Y.; and Haffner, P.
\newblock 1998.
\newblock Gradient-based learning applied to document recognition.
\newblock {\em Proceedings of the IEEE} 86(11):2278--2324.

\bibitem[\protect\citeauthoryear{Leen and Orr}{1994}]{ann94}
Leen, T.~K., and Orr, G.~B.
\newblock 1994.
\newblock Optimal stochastic search and adaptive momentum.
\newblock In {\em Advances in neural information processing systems},
  477--484.

\bibitem[\protect\citeauthoryear{Li and Long}{2000}]{rel00}
Li, Y., and Long, P.~M.
\newblock 2000.
\newblock The relaxed online maximum margin algorithm.
\newblock In {\em Advances in neural information processing systems},
  498--504.

\bibitem[\protect\citeauthoryear{Mahsereci \bgroup et al\mbox.\egroup
  }{2017}]{ear17}
Mahsereci, M.; Balles, L.; Lassner, C.; and Hennig, P.
\newblock 2017.
\newblock Early stopping without a validation set.
\newblock {\em arXiv preprint arXiv:1703.09580}.

\bibitem[\protect\citeauthoryear{Nettleton, Orriols-Puig, and
  Fornells}{2010}]{net10}
Nettleton, D.~F.; Orriols-Puig, A.; and Fornells, A.
\newblock 2010.
\newblock A study of the effect of different types of noise on the precision of
  supervised learning techniques.
\newblock {\em Artificial intelligence review} 33(4):275--306.

\bibitem[\protect\citeauthoryear{Patrini \bgroup et al\mbox.\egroup
  }{2017}]{lca17}
Patrini, G.; Rozza, A.; Menon, A.~K.; Nock, R.; and Qu, L.
\newblock 2017.
\newblock Making deep neural networks robust to label noise: a loss correction
  approach.
\newblock {\em stat} 1050:22.

\bibitem[\protect\citeauthoryear{Paudice, Mu{\~n}oz-Gonz{\'a}lez, and
  Lupu}{2018}]{poi18}
Paudice, A.; Mu{\~n}oz-Gonz{\'a}lez, L.; and Lupu, E.~C.
\newblock 2018.
\newblock Label sanitization against label flipping poisoning attacks.
\newblock {\em arXiv preprint arXiv:1803.00992}.

\bibitem[\protect\citeauthoryear{Reed \bgroup et al\mbox.\egroup
  }{2014}]{boo14}
Reed, S.; Lee, H.; Anguelov, D.; Szegedy, C.; Erhan, D.; and Rabinovich, A.
\newblock 2014.
\newblock Training deep neural networks on noisy labels with bootstrapping.
\newblock {\em arXiv preprint arXiv:1412.6596}.

\bibitem[\protect\citeauthoryear{Rolnick \bgroup et al\mbox.\egroup
  }{2017}]{rol17}
Rolnick, D.; Veit, A.; Belongie, S.; and Shavit, N.
\newblock 2017.
\newblock Deep learning is robust to massive label noise.
\newblock {\em arXiv preprint arXiv:1705.10694}.

\bibitem[\protect\citeauthoryear{Sukhbaatar \bgroup et al\mbox.\egroup
  }{2014}]{tcn14}
Sukhbaatar, S.; Bruna, J.; Paluri, M.; Bourdev, L.; and Fergus, R.
\newblock 2014.
\newblock Training convolutional networks with noisy labels.
\newblock {\em arXiv preprint arXiv:1406.2080}.

\bibitem[\protect\citeauthoryear{Tarvainen and Valpola}{2017}]{mea17}
Tarvainen, A., and Valpola, H.
\newblock 2017.
\newblock Mean teachers are better role models: Weight-averaged consistency
  targets improve semi-supervised deep learning results.
\newblock In {\em Advances in neural information processing systems},
  1195--1204.

\bibitem[\protect\citeauthoryear{Vahdat}{2017}]{crf17}
Vahdat, A.
\newblock 2017.
\newblock Toward robustness against label noise in training deep discriminative
  neural networks.
\newblock In {\em Advances in Neural Information Processing Systems},
  5601--5610.

\bibitem[\protect\citeauthoryear{Van~Rooyen, Menon, and
  Williamson}{2015}]{unh15}
Van~Rooyen, B.; Menon, A.; and Williamson, R.~C.
\newblock 2015.
\newblock Learning with symmetric label noise: The importance of being
  unhinged.
\newblock In {\em Advances in Neural Information Processing Systems},  10--18.

\bibitem[\protect\citeauthoryear{Xiao \bgroup et al\mbox.\egroup
  }{2015}]{lfm15}
Xiao, T.; Xia, T.; Yang, Y.; Huang, C.; and Wang, X.
\newblock 2015.
\newblock Learning from massive noisy labeled data for image classification.
\newblock In {\em Proceedings of the IEEE Conference on Computer Vision and
  Pattern Recognition},  2691--2699.

\bibitem[\protect\citeauthoryear{Xiao, Rasul, and Vollgraf}{2017}]{fmn17}
Xiao, H.; Rasul, K.; and Vollgraf, R.
\newblock 2017.
\newblock Fashion-mnist: a novel image dataset for benchmarking machine
  learning algorithms.
\newblock {\em arXiv preprint arXiv:1708.07747}.

\bibitem[\protect\citeauthoryear{Zagoruyko and Komodakis}{2016}]{wrn16}
Zagoruyko, S., and Komodakis, N.
\newblock 2016.
\newblock Wide residual networks.
\newblock {\em arXiv preprint arXiv:1605.07146}.

\bibitem[\protect\citeauthoryear{Zhu, Wu, and Chen}{2003}]{eli03}
Zhu, X.; Wu, X.; and Chen, Q.
\newblock 2003.
\newblock Eliminating class noise in large datasets.
\newblock In {\em Proceedings of the 20th International Conference on Machine
  Learning (ICML-03)},  920--927.

\end{thebibliography}

\end{document}